\documentclass[conference]{IEEEtran}
\usepackage[utf8]{inputenc}
\usepackage[english]{babel}

\usepackage{comment}
\usepackage{xspace}
\usepackage{graphicx}
\usepackage{overpic}
\usepackage{svg}
\usepackage{xcolor}
\usepackage{bm}
\usepackage{tikz-cd}
\usepackage[hang,flushmargin]{footmisc}
\usepackage{adjustbox}
\usepackage{etoolbox}
\usepackage{environ}
\usepackage{pbalance}
\usepackage{flushend}
\usepackage{mathtools}
\usepackage[normalem]{ulem}

\usepackage{tabularx}
\usepackage{multirow}
\usepackage{booktabs}
\usepackage{colortbl}
\usepackage{float}
\usepackage{stfloats}   %
\usepackage{placeins}
\usepackage{array}
\usepackage{threeparttable}

\usepackage{amsmath}
\usepackage{amssymb}
\usepackage{amsthm}
\usepackage{siunitx}

\DeclareMathOperator*{\argmax}{arg\,max}
\DeclareMathOperator*{\argmin}{arg\,min}

\newcommand{\ie}{\emph{i.e.}\@\xspace}

\newcommand{\vs}{\emph{vs.}\@\xspace}

\makeatletter
\patchcmd{\@makecaption}{\scshape}{}{}{}
\newcommand{\onetagright}{\tagsleft@false}
\makeatother

\usepackage[font=footnotesize,labelsep=period]{caption}
\usepackage{subcaption}
\newtheoremstyle{main}
{1em}                                                %
{1em}                                                %
{\itshape}                                           %
{0pt}                                                %
{\scshape}                                           %
{\\*}                                                %
{2pt}                                                %
{\thmname{#1}\thmnumber{ #2}: \thmnote{\itshape #3}} %

\usepackage{algorithm}
\usepackage{algpseudocode}
\newcommand{\removelatexerror}{\let\@latex@error\@gobble}

\usepackage[inline]{enumitem}

\usepackage[
	activate   = {true},
	protrusion = false,
	expansion  = true,
	kerning    = true,
	spacing    = true,
	tracking   = false,
	auto       = true,
	selected   = true,
	factor     = 1000,
	stretch    = 10,
	shrink     = 10,
]{microtype}

\usepackage{csquotes}
\usepackage[
	maxbibnames=4, minbibnames=1,
	maxcitenames=2,
	natbib=true,
	bibstyle=ieee,
	citestyle=numeric-comp,
	backend=biber,
	sorting=none,
	giveninits=true,
	url=false,
	doi=false,
	eprint=false,
	isbn=false,
]{biblatex}

\definecolor{purduegold}{HTML}{C28E0E} %

\makeatletter
\let\NAT@parse\undefined
\makeatother
\usepackage[pdfa,colorlinks,bookmarksopen,bookmarksnumbered,allcolors=purduegold]{hyperref}
\usepackage{bookmark}

\usepackage[nameinlink,capitalise]{cleveref}
\crefname{line}{line}{lines}
\crefname{figure}{Fig.}{Figs.}
\Crefname{figure}{Fig.}{Figs.}
\crefname{equation}{Eq.}{Eqs.}
\Crefname{equation}{Eq.}{Eqs.}
\crefname{section}{Sec.}{Secs.}
\Crefname{section}{Sec.}{Secs.}
\crefname{definition}{Def.}{Defs.}
\Crefname{definition}{Def.}{Defs.}
\crefname{algorithm}{Alg.}{Algs.}
\Crefname{algorithm}{Alg.}{Algs.}
\crefname{assumption}{Asm.}{Asms.}
\Crefname{assumption}{Asm.}{Asms.}
\crefname{subassumption}{Asm.}{Asms.}
\Crefname{subassumption}{Asm.}{Asms.}
\crefname{problem}{Problem}{Problems}
\Crefname{problem}{Problem}{Problems}

\makeatletter
\newcommand\footnoteref[1]{\protected@xdef\@thefnmark{\ref{#1}}\@footnotemark}
\makeatother

\title{Sampling-based Model Predictive Control \\ as Sequential Monte Carlo Inference}
\IEEEoverridecommandlockouts
\author{Austin Garrett, Muyang Yan, and Zachary Kingston%
  \thanks{This work has been submitted to the IEEE for possible publication. Copyright may be transferred without notice, after which this version may no longer be accessible.}%
  \thanks{Department of Computer Science, Purdue University, USA.}%
}

\makeatletter
\apptocmd{\@maketitle}{%
  \vspace{2ex}%
  \centering
  \includegraphics[width=\textwidth]{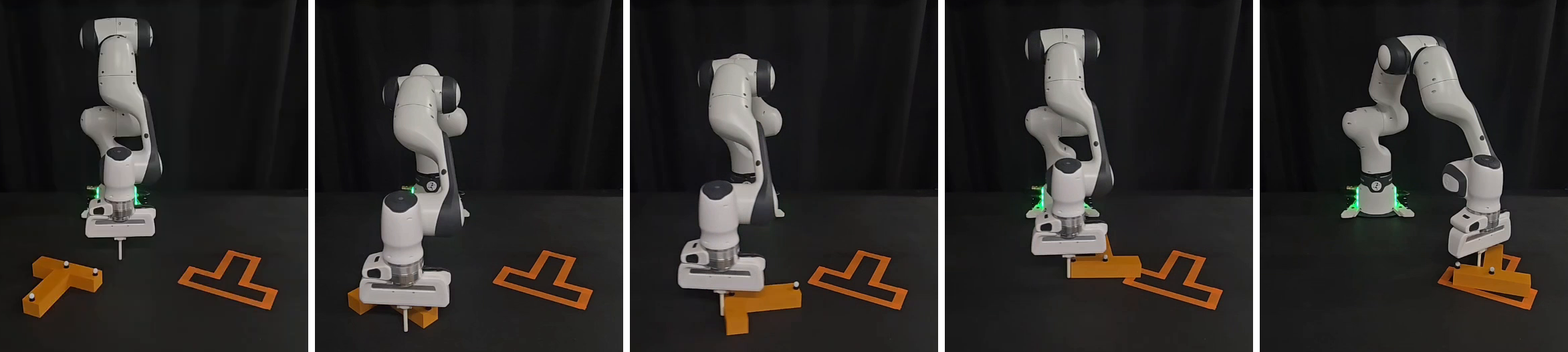}%
  \setcounter{figure}{0}%
  \captionof{figure}{Non-prehensile Push-T solved by training-free SMC on a Franka FR3. The method replans throughout the 30 second trial, from the initial pose to the goal pose. Our programmable proposal reaches 90\% success over 20 randomized start and goal poses, against 10--15\% for the baselines (\cref{sec:franka}).
  }%
  \label{fig:fr3-setup}%
  \vspace{-1ex}%
}{}{}
\makeatother

\begin{document}

\maketitle

\begin{abstract}
  Sampling-based model predictive control (MPC) is treated as a family of distinct algorithms: predictive sampling, MPPI, the cross-entropy method, and their biased, colored-noise, and annealed variants all sample control sequences and reweight them by cost.
  We show a single Sequential Monte Carlo (SMC) inference algorithm implements them all, specified by three independent axes: the proposal, the update rule that reduces cost-weighted particles to a nominal plan, and the effective-sample-size threshold for resampling.
  We evaluate SMC-based planning on MuJoCo non-linear control, open-loop navigation through obstacle fields, 7-DoF motion planning, and non-prehensile push-T, in simulation and on a Franka FR3.
  Resampling combats particle degeneracy at long horizons.
  At a 28-knot horizon in our navigation domain it cut the collision rate by 30\% over non-resampling baselines.
  Programmable proposals improve push-T success over perturbative baselines, improving from 1\% to 100\% in simulation and 10\% to 90\% on hardware.
\end{abstract}

\section{Introduction}

Model Predictive Control (MPC) optimizes a sequence of actions over a finite horizon subject to a dynamics model and has seen great success in autonomous racing~\cite{williams2016driving}, humanoid navigation~\cite{dialmpc}, and dynamic manipulation~\cite{storm}.
Among MPC methods, sampling-based approaches make the trade-off between compute and trajectory quality explicit through the number of rollouts.
They are also gradient-free, requiring only forward rollouts, and suit non-linear dynamics and non-convex costs, which confound gradient-based optimization~\cite{suh2022differentiable, pang2023global}.
However, sampling-based methods face two related challenges as planning horizons grow: (1) they are often based on Gaussian perturbations which cause multi-modal cost landscapes to collapse into a single mode, and (2) their importance weights degenerate.

Both challenges trace to the sampling distribution over controls.
Reinforcement learning can learn the distribution as a task-specific proposal, but requires training for every new task.
Learning-free MPC instead perturbs a nominal control sequence with Gaussian noise, as in predictive sampling, MPPI~\cite{williams2017mppi}, and CEM~\cite{rubinstein2004cem}.
Sampling-based MPC also updates every knot at once, which exploits temporal parallelism but ignores the sequential structure of the trajectory, and thus performance degrades as the horizon grows.
We address both challenges by casting MPC as approximate posterior inference in a generative model whose likelihood is an exponentiated negative cost~\cite{rl-and-control-as-inference}.
This framing separates the dynamics, proposal, and cost, each of which can be learned or programmed and then combined at inference time.

Sequential Monte Carlo (SMC)~\cite{doucet2011tutorial} suits trajectory optimization: it interleaves proposal sampling, likelihood weighting, and resampling to refine a particle approximation of the target distribution.
It has been applied to reinforcement learning and planning~\cite{lazaric2007reinforcement, piche2018probabilistic}, autonomous driving~\cite{lioutas2023critic}, and character animation~\cite{hamalainen2014online}.
Its expressivity is also a challenge: how proposals, schedules, and resampling strategies interact in MPC is not well understood.
Thus, we contribute:
\begin{enumerate}
  \item A particle decomposition of sampling-based MPC that recovers Predictive Sampling, MPPI, CEM, and their variants by ablating the proposal, update, and resampling threshold.
    Earlier unifications propagate a distribution parameter~\cite{okada2019vimpc, wang2021tsallis, watson2022inferring, wagener2019dmdmpc}; we propagate a weighted particle set which adds resampling inside the horizon and admits proposals with no weighted maximum-likelihood projection.
  \item A study of the three axes across four domains including MuJoCo simulation, collision-free motion planning, and a push-T task both in simulation and on a real robot.
  \item An open-source\footnote{\url{https://github.com/CoMMALab/PFMP/}} vectorized JAX implementation that lifts the particle-count limits of prior SMC planners~\cite{piche2018probabilistic} to run in real time on a GPU and allows users to write programmable proposals.
\end{enumerate}

\section{Related Work}

Our approach builds on the literature casting control and planning as probabilistic inference.
The foundational insight is that optimal control reduces to inference in a graphical model whose likelihood encodes cost~\cite{todorov2007linearly, kappen2012optimal, rl-and-control-as-inference}.
Within this framework trajectory optimization becomes approximate posterior inference.
SMC is natural here given the sequential structure of trajectory distributions: it originated as particle filtering for nonlinear state estimation~\cite{gordon1993novel}, matured into a general methodology for sequential inference~\cite{doucet2001smc, doucet2011tutorial}, and in its SMC samplers form~\cite{delmoral2006smcsamplers} recovers annealed importance sampling~\cite{neal2001annealed} as the special case of a particular backward kernel.

Several works apply SMC directly to planning.
\citet{piche2018probabilistic} cast planning as density estimation over optimal future trajectories, using a learned policy as the proposal with weights from a maximum-entropy advantage function; the method needs many particles and assumes a jointly learned dynamics model and value function.
Our planner instead operates at inference time from a programmable dynamics model and cost.
\citet{kantas2009smc} use SMC as a global optimizer for non-convex stochastic MPC via annealing, but their particle distribution collapses to a Dirac on the optimum, whereas we retain a weighted approximation of the full controlled distribution.
SMC has also been used for various adjacent objectives such as multimodal goal inference~\cite{zhixuan2020sips}, physics-based character animation~\cite{hamalainen2014online}, and autonomous driving with learned critics~\cite{lioutas2023critic}.
A recent extension called twisted SMC~\cite{zhao2024twisted} explores a richer design space for intermediate target distributions.
Similar concepts to resampling have been used in sampling-based MPC.
\citet{yin2025safebeyond} resample rollouts that violate a state constraint at every step, and \citet{yang2026tsmc} resamples along a tempering schedule, with a fixed resampling rule.
Our resampling rule is instead a tunable parameter which we vary in our experiments.

A large family of sampling-based MPC methods generate candidate control sequences and reweight them by cost.
Closest to ours is Model Predictive Path Integral control (MPPI)~\cite{williams2017mppi}, and its recent variant Biased-MPPI~\cite{trevisan2024biasedmppi}, which fuses ancillary controllers into the sampling distribution.
MPPI built upon earlier methods such as STOMP, which introduced cost-weighted updates~\cite{kalakrishnan2011stomp}, and it inspired variants that parallelize sampling~\cite{pezzato2023sampling}, guide it via an unscented transform~\cite{mohamed2024umppi}, derive an optimal sampling covariance~\cite{yi2024covompc}, or learn sampling distributions directly~\cite{sacks2022learning, power2022variational}.
The Cross-Entropy Method (CEM)~\cite{rubinstein2004cem} and its temporally-correlated variant iCEM~\cite{pinneri2021icem} are also presented as separate algorithms; they differ from MPPI only in proposal, resampling threshold, and update rule (\cref{tab:decomposition}).
Structured proposals are a related line of work: Model Tensor Planning~\cite{modeltensorplanning} draws control sequences over randomized multipartite graphs and mixes them with local perturbations.
Through our ablations we find that among these extensions, the ancillary controller (a proposal, in SMC terms), not the cost-weighted mean update, is critical for harder tasks.

Sampling-based MPC has seen other recent unification attempts; \citet{honda2026pimpc} surveys the design space and reduces it to two elements, the prior and the variational family.
\citet{okada2019vimpc} recover CEM, MPPI, and CMA-ES as variational inference under different choices of divergence and trajectory distribution, and \citet{wang2021tsallis} generalize to also cover Stein variational MPC.
\citet{watson2022inferring} organize MPPI, CEM, and PI\textsuperscript{2} as Gibbs posteriors over a prior on action sequences.
Each of these propagates a distribution parameter between iterations and projects its samples back into a parametric family.
We propagate a weighted particle set and its ancestry instead, which adds resampling inside the horizon as a third independent axis.

More recently, other generative methods have arisen in parallel, though often through fusing learned value functions to their controller, and without decomposing their algorithm into independent axes that we specify as typed components.
DMD-MPC~\cite{wagener2019dmdmpc} casts MPC as online learning, separating its axes into choices of objective, control distribution, and Bregman divergence, which recovers MPPI exactly and CEM up to a moment-matrix substitution. Those choices span our update and proposal axes; its instantiations stay in the exponential family, and without importance weights it does not admit a resampling layer.

Two recent generative methods sit adjacent to this work.
Model-Based Diffusion~\cite{pan2024modelbased} computes a diffusion score directly from dynamics rather than from data and also carries no importance weights and no resampling layer so it is not included in our decomposition.
Generative Predictive Control~\cite{kurtz2025generativepredictivecontrolflow} follows in that data-free vein, training a flow-matching policy on planner-generated roll-outs and sampling from it online. This is a learned proposal, similar to \citeauthor{piche2018probabilistic}'s policy, in that it requires an offline training phase.
In contrast, our method forward-propagates importance weights in a particle-based algorithm.
Like other recent programmable~\cite{cusumanotowner2019gen} and vectorized probabilistic programming systems~\cite{becker2026vectorized}, we target the GPU for efficient inference.

\section{Model Predictive Control}

Let \(\mathcal{X}\) be a state space and \(\mathcal{U}\) a control space, neither of which need be a vector space.
Model Predictive Control (MPC) is constrained by a deterministic state transition map:
\[
  D: \mathcal{X} \times \mathcal{U} \to \mathcal{X},\ \ \ \ (x_t,u_t) \mapsto x_{t+1}.
\]

Let \(\Theta\) be a named record type (like a typed tuple).
Let \(\theta \in \Theta\) be a \textit{domain parameter} and let \(D := D_\theta\) be the domain-parameterized transition.
The choice of domain parameter induces an instantaneous cost on states and controls:
\[
  C(\cdot\ ;\ \theta) : \mathcal{X} \times \mathcal{U} \to \mathbb{R}^+,\ \ \ \ (x_t, u_t) \mapsto C_t,
\]
where \(C_t \geq 0\) for all times \(t \in [0\cdots H]\).
The cost of a full sequence \([u_0\cdots u_{H}] = \mathbf{u} \in \mathcal{U}^{H+1}\) unfolds the parameterized transition \(H\) times from an initial state \(x_0 \in \mathcal{X}\):
\[
  \begin{tikzcd}[column sep=small]
    {(x_0, u_0)} & \cdots & {(x_{H-1},u_{H-1})} & {x_{H}} \\
    {C_0} && {C_{H-1}} & {C_\mathrm{term}}
    \arrow["{D_\theta}", from=1-1, to=1-2]
    \arrow["{C(\cdot\ ;\ \theta)}", from=1-1, to=2-1]
    \arrow["{D_\theta}", from=1-2, to=1-3]
    \arrow["{C(\cdot\ ;\ \theta)}", from=1-3, to=2-3]
    \arrow["{D_\theta}", from=1-3, to=1-4]
    \arrow["{C_\mathrm{term}(\cdot\ ;\ \theta)}", from=1-4, to=2-4]
\end{tikzcd}\]
where
\[
  S(x_0,\mathbf{u}; \theta) = C_\mathrm{term}(x_H; \theta) + \sum_{t=0}^{H-1} C(x_t,u_t; \theta)
\]
is the sum of a terminal cost and running costs over a finite horizon.
The optimal objective of MPC is the lowest-cost command sequence \([u^\star_0\ldots u^\star_{H}] = \mathbf{u}^\star \in \mathcal{U}^{H+1}\), \ie,
\[
  \mathbf{u}^* = \argmin_{\mathbf{u}\ \in\ \mathcal{U}^{H+1}}\ S(x_0,\mathbf{u}; \theta).
\]

MPC runs in a closed-loop with a receding horizon.
After computing the optimal sequence, the first action is executed and the horizon is advanced.

\subsection{Model Predictive Control as Approximate Inference}

MPC can be cast as approximate probabilistic inference with a cost function.
Following \citet{rl-and-control-as-inference}, we introduce an auxiliary random variable that recasts the cost as a conditional probability.
Let \(\mathcal{O} \in \{0, 1\}\) be the \textit{optimality indicator}.
The optimality likelihood is the exponentiated negative cost:
\begin{align}
  p(\mathcal{O} = 1 \mid \mathbf{u}, x_0;\,\theta) = \exp(-S(x_0,\mathbf{u}; \theta)).
\end{align}

\noindent The initial controls are distributed according to a prior \(q(\mathbf{u} \mid x_0;\,\theta)\), called the \textit{uncontrolled distribution}.
Using Bayes' rule, the target \textit{controlled distribution} is proportional to the probability of a sequence under the uncontrolled prior distribution, multiplied by the exponentiated negative cost:
\begin{align}
  p(\mathbf{u} \mid \mathcal{O} = 1, x_0;\,\theta) \propto q(\mathbf{u} \mid x_0;\,\theta) \cdot \exp(-S(x_0, \mathbf{u}; \theta)).
\end{align}
\noindent This distribution spreads mass over many trajectories, weighting softly-optimal solutions by cost.
Under a uniform path prior \(q(\mathbf{u} \mid x_0;\,\theta) = \mathbb{U}[\mathcal{U}]\), the hard MPC objective above is the maximum a-posteriori (MAP) objective of the controlled distribution:
\[
  \mathbf{u}^* = \argmin_{\mathbf{u}\ \in\ \mathcal{U}^{H+1}}\ S(x_0,\mathbf{u}; \theta) = \argmax_{\mathbf{u}\ \in\ \mathcal{U}^{H+1}}\ p(\mathbf{u} \mid \mathcal{O} = 1, x_0;\,\theta).
\]

While classical MPC is recovered as the MAP of the controlled distribution, sampling-based methods such as MPPI~\cite{williams2016driving} operate on the controlled distribution directly, using posterior expectations to compute a policy.
The \textit{expected controlled trajectory} is
\begin{align*}
  \mathbb{E}[\mathbf{u} \mid \mathcal{O} = 1, x_0;\,\theta]
  &= \frac{\int \mathbf{u}\, q(\mathbf{u} \mid x_0;\,\theta)\, \exp(-S)\ d\mathbf{u}}
  {\int q(\mathbf{u} \mid x_0;\,\theta)\, \exp(-S)\ d\mathbf{u}},
\end{align*}
where \(S = S(x_0, \mathbf{u}; \theta)\).
Under an uncontrolled prior, the denominator here is an intractable path integral over \(\mathcal{U}^{H+1}\).
This motivates a sequential factorization of the posterior.

\section{Sequential Monte Carlo for MPC}

The target controlled distribution decomposes into distributions \([\pi_{t'}]_{t'=0}^H\) approximated by particles.
Let the uncontrolled prior distribution be a Markov chain
\[
  q(\mathbf{u} \mid x_0;\,\theta) = q_0(u_0\ |\ x_0;\ \theta)\prod_{t=1}^{H} q_t(u_t\ |\ u_{t-1}, x_{t-1};\ \theta),
\]
consisting of two state-conditioned samplers \((q_0, q_t)\), called the \textit{initial} control proposal and the \textit{running} control proposal.
For every \(t' \in [0,\dots, H]\), the exponentiated negative path cost factorizes into a product:
\[
  \exp\left(-\sum_{t=0}^{t'} C_t\right) = \prod_{t=0}^{t'} \exp (-C_t),
\]
so that the target partial controlled distributions are
\begin{equation*}
  \begin{split}
    \pi_{t'} \propto{}\ & q_0(u_0 \mid x_0;\,\theta) \exp(-C_0) \\
    & \cdot \prod_{t=1}^{t'} q_t(u_t \mid u_{t-1}, x_{t-1};\,\theta) \exp(-C_t),
  \end{split}
\end{equation*}
with the final controlled distribution at \(t' = H\) additionally weighted by the terminal cost \(\exp(-C_\mathrm{term}(x_H; \theta))\).

Sequential Importance Sampling with Resampling (SIS/R) is the standard practical instantiation of SMC~\cite{doucet2001smc, doucet2011tutorial} which is a scheme for propagating a weighted particle set through a sequence of intermediate target distributions. SIS/R specifically uses an \textit{importance weighting} scheme to extend and re-weight each particle, and a resampling step to compensate for the weight degeneracy SIS encounters as sequences grow longer.
Our planner maintains a population of particles and extends them over the finite horizon, accumulating cost over the lifetime of each particle, and resamples when diversity, measured by the \textit{effective sample size}~\cite{kong1994sequential}, drops below a critical threshold~\cite{liu1998sequential} $\rho \in [0, 1]$.
SIS corresponds to $\rho = 0$, while SIS/R is $\rho \in (0, 1]$.

\begin{algorithm}[t]
  \caption{Sequential Importance Sampling-Resampling}
  \label{alg:sisr}
  \begin{algorithmic}[1]

    \Require \(x_0,\; \theta,\; H,\; K,\; \rho\)
    \Ensure Nominal plan \(\mu \in \mathcal{U}^{H+1}\)

    \State Sample \(u_0^{(i)} \sim q_0(\cdot \mid x_0;\,\theta)\),\;
    \(\tilde{w}_0^{(i)} \gets 0\) \quad for \(i \in [K]\)

    \medskip

    \For{\(t = 1, \ldots, H\)}
    \For{\(i = 1, \ldots, K\)} \Comment{Propose}
    \State \(\nu_t^{(i)} \sim q_t(\cdot \mid u_{t-1}^{(i)}, x_{t-1}^{(i)};\, \theta)\)
    \State \(x_t^{(i)} \gets D_\theta(x_{t-1}^{(i)},\, \nu_t^{(i)})\)
    \State \(\tilde{\omega}_t^{(i)} \gets \tilde{w}_{t-1}^{(i)}
    - C(x_t^{(i)},\, \nu_t^{(i)};\, \theta)\)
    \EndFor

    \medskip
    \State \(p_t \gets \mathrm{softmax}(\tilde{\omega}_t^{(1:K)})\) \Comment{Resample}
    \If{\(\mathrm{ESS}(p_t) < \rho K\)}
    \State \(a_t^{(i)} \sim \mathrm{Cat}(p_t^{(1:K)})\), \quad \(\tilde{w}_t^{(i)} \gets 0\)
    \Else
    \State \(a_t^{(i)} \gets i\), \quad \(\tilde{w}_t^{(i)} \gets \tilde{\omega}_t^{(i)}\)
    \EndIf

    \State \(u_t^{(i)} \gets \nu_t^{(a_t^{(i)})}\),\;\;
    \(x_t^{(i)} \gets x_t^{(a_t^{(i)})}\)
    \quad for \(i \in [K]\)

    \EndFor

    \medskip
    \State \(\tilde{w}_\mathrm{final}^{(i)} \gets \tilde{w}_{H}^{(i)}
    - C_{\mathrm{term}}(x_H^{(i)};\theta)\)
    \quad for \(i \in [K]\) \Comment{Terminal}

    \medskip
    \State Trace ancestors \(\{a_t\}\) to recover \(\mathbf{u}^{(i)}\)
    \State \(\mu \gets U(\mathbf{u}^{(i)},\, \tilde{w}_\mathrm{final}^{(i)})\) \Comment{Update (\cref{tab:decomposition})}
    \State \textbf{Execute} \(\mu_0\); shift horizon
  \end{algorithmic}
\end{algorithm}

\begin{figure*}[t]
  \centering
  \includegraphics[width=\textwidth]{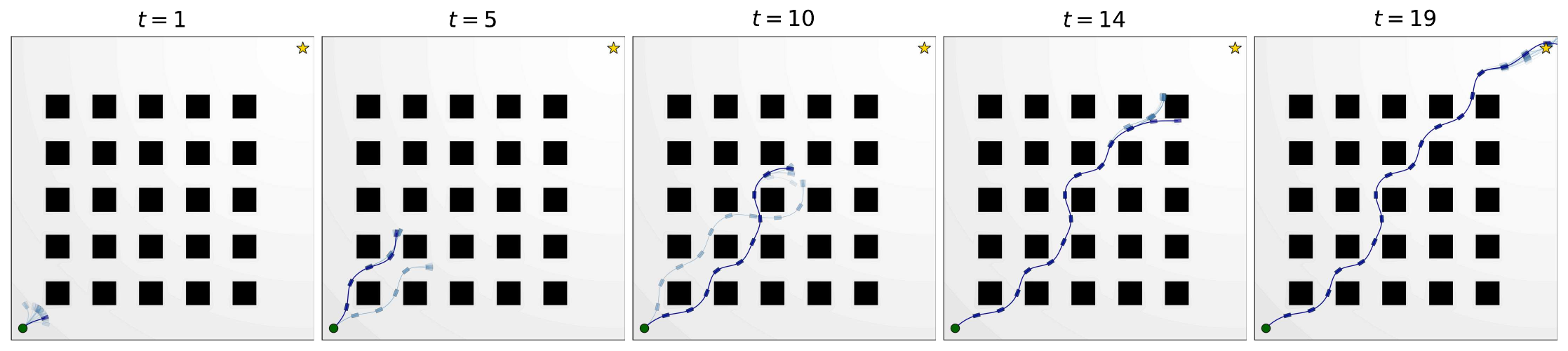}
  \caption{The sequential extension on a \(5\times5\) obstacle grid, from the start (green) to the goal (star).
    Each panel is a slice \(t\) of one rollout: \(K = 1024\) particles are extended by one knot, reweighted by the accumulated cost, and resampled whenever the normalized effective sample size falls below \(\rho = 0.5\).
    The twenty lowest-cost trajectories are drawn with opacity by rank, darkest best.
  Early on the population fans out across several routes through the grid, and over time it increasingly concentrates on a successful mode.}
  \label{fig:filmstrip}
\end{figure*}

\cref{alg:sisr} summarizes the complete SIS/R procedure.
\cref{fig:filmstrip} shows the extension and adaptive pruning it performs, one knot at a time, on a regular obstacle grid.
In practice, the running cost \(C\) in the \textsc{Propose} step is accumulated by numerically integrating over sub-steps within each interval: the proposal \(q\) emits a new knot, and the dynamics \(D_\theta\) are evaluated over the resulting interpolated trajectory segment.

\subsection{A Unifying Decomposition}

Our control task is abstractly captured by two variables: the domain \(\theta \in \Theta\), a record type aggregating all data held constant during a single planning window, and the dynamics \(D_\theta : \mathcal{X} \times \mathcal{U} \to \mathcal{X}\), the deterministic, parametric model of how actions affect states.
Given a task, SMC forms a sampling-based MPC algorithm from three independent, typed axes.
\begin{enumerate}
  \item \textbf{Proposal} \(q : \mathcal{X} \times \Theta \to \Delta(\mathcal{U}^{H+1})\): the Markov chain from which candidate action sequences are generated. \(\Delta(\mathcal{U}^{H+1})\) denotes distributions over \(\mathcal{U}^{H+1}\).
  \item \textbf{Resampling threshold} \(\rho \in [0,1]\): the effective-sample-size fraction below which we resample the self-normalized ensemble weights (SIS \vs SIS/R).
  \item \textbf{Update rule} \(U : (\mathcal{U}^{H+1} \times \mathbb{R})^K \to \mathcal{U}^{H+1}\): reduces the \(K\) cost-weighted samples to the nominal plan \(\mu\).
\end{enumerate}

Several well-known MPC algorithms fundamentally compute the exact same weights as the \textit{Sequential Importance Sampling} (SIS) algorithm. We summarize this connection in \cref{tab:decomposition}.

\begin{table}[b]
  \vspace{-1em}
  \centering
  \caption{SMC-MPC: Sequential Monte Carlo decomposition of Model Predictive Control into three independent axes}
  \label{tab:decomposition}
  \begin{tabular}{lccc}
    \toprule
    \textbf{MPC Algorithm} & \textbf{Proposal} & \textbf{Resample} (\(\rho\)) & \textbf{Update} \\
    \midrule
    PS~\cite{howell2022predictive} & \(q_{\mathrm{perturb}}\) & \(= 0\) & \(U_{\mathrm{MAP}}\) \\
    MPPI~\cite{williams2017mppi} & \(q_{\mathrm{perturb}}\) & \(= 0\) & \(U_{\mathrm{mean}}\) \\
    Biased-MPPI~\cite{trevisan2024biasedmppi} & Any & \(= 0\) & \(U_{\mathrm{mean}}\) \\
    DIAL-MPC~\cite{dialmpc} & \(q_{\mathrm{perturb}}\) & \(= 0\) & \(U_{\mathrm{DIAL}}\) \\
    CEM~\cite{rubinstein2004cem} & \(q_{\mathrm{perturb}}\) & \(= 0\) & \(U_{\mathrm{elite}}\) \\
    iCEM~\cite{pinneri2021icem} & \(q_{\mathrm{colored}}\) & \(= 0\) & \(U_{\mathrm{elite}}\) \\
    \midrule
    \citet{piche2018probabilistic} & \(q_{\mathrm{learned}}\) & \(= 1\) & \(U_{\mathrm{sample}}\) \\
    SMC-MPC (ours)  & Any & \(\in [0,1]\) & Any \\
    \bottomrule
  \end{tabular}
\end{table}

Resampling (SIS/R) adds an additional degree of freedom. At each step, the ancestor indices \(a_t^{(i)}\) record which parent each particle was cloned from.
A single particle index \(i\) at time \(H\) is not one trajectory; its path is a lineage through the tree given by the ancestor chain \(a_H^{(i)}, a_{H-1}^{(a_H^{(i)})}, \ldots\)\,.
Before applying the update rule \(U\), we recover full trajectories by walking this chain backward from \(t = H\) to \(t = 0\).
This is standard in particle smoothing~\cite{doucet2011tutorial} and runs in \(O(HK)\) time.

Given the reconstructed candidate action sequences \(\mathbf{u}^{(i)}\) and their normalized cost weights \(\tilde{w}^{(i)}\), the update rule returns a nominal plan:
\[
  \mu = U\!\left(\{(\mathbf{u}^{(i)},\, \tilde{w}^{(i)})\}_{i=1}^K\right).
\]

We allow \(\mu\) to be a sub-component of \(\theta\), as it is held constant over the finite horizon.
After planning, the first control \(\mu_0\) of the nominal plan is executed on the plant and the horizon is shifted forward by \(\delta t\), determined as part of the dynamics.

\subsubsection{Recovering MPC Baselines}
\label{sec:baselines}
A common proposal in this setting is to sample Gaussian perturbations around the current nominal plan, assuming that the control at each step in time is an independent vector in \(\mathbb{R}^{n_u}\):
\[
  q_{\mathrm{perturb}}(\cdot \mid u_{t-1}^{(i)}, x_{t-1}^{(i)}; \theta) = \mathcal{N}(\mu_t,\; \mathrm{diag}(\sigma_1, \ldots, \sigma_{n_u})).
\]

Under the perturbation proposal and when \(\rho = 0\), the outer closed-loop update rules correspond exactly to the update equations of several common MPC algorithms.
Setting \(\mu\) to the trajectory with highest weight gives \(U_{\mathrm{MAP}}\), which recovers \textbf{Predictive Sampling} (PS):
\[
  \mu_{\mathrm{MAP}} = \mathbf{u}^{(\arg\max_i \tilde{w}^{(i)})}.
\]

To distinguish it from its Gaussian counterpart, when PS comes with a programmable proposal, we refer to it as \textbf{Biased-PS}.
The weights may be tempered, \(\tilde{w}^{(i)} \propto \exp(-S^{(i)}/\lambda)\).
Temperature \(\lambda\) rescales the cost-likelihood and is absorbed into \(C(\cdot;\theta)\); our SMC target fixes \(\lambda=1\), while MPPI's \(\lambda\) is a free hyperparameter tuned per task.
CEM has no temperature: it ranks its elite set, so it is invariant to monotone rescaling of cost.
The expected trajectory under the cost-weighted distribution gives \(U_{\mathrm{mean}}\), which recovers \textbf{MPPI}:
\[
  \mu_{\mathrm{mean}} = \sum_i \tilde{w}^{(i)} \mathbf{u}^{(i)}.
\]

Retaining the \(U_{\mathrm{mean}}\) update but replacing the perturbation proposal with a custom proposal \(q_{\mathrm{custom}}\) that biases samples toward an ancillary controller yields \textbf{Biased-MPPI}~\cite{trevisan2024biasedmppi}, illustrating that the proposal and update axes vary independently.

If \(\mu\) is set to the unweighted mean, and \(\sigma\) to the unweighted standard deviation of an elite subset \(\mathcal{E} = \{i \mid \mathrm{rank}(\tilde{w}^{(i)}) \leq K_e\}\) for elite count \(K_e \leq K\), the resulting \(U_{\mathrm{elite}}\) recovers the \textbf{Cross-Entropy Method} (CEM):
\[
  \mu_{\mathrm{elite}} = \frac{1}{|\mathcal{E}|} \sum_{i \in \mathcal{E}} \mathbf{u}^{(i)}, \quad
  \sigma_{\mathrm{elite}} = \sqrt{\frac{1}{|\mathcal{E}|} \sum_{i \in \mathcal{E}} \left(\mathbf{u}^{(i)} - \mu_{\mathrm{elite}}\right)^2}.
\]

Likewise, retaining the elite update while replacing the perturbation proposal yields \textbf{iCEM}~\cite{pinneri2021icem} under temporally-correlated colored noise \(q_{\mathrm{colored}}\), and \textbf{Biased-CEM} under \(q_{\mathrm{custom}}\) (no published counterpart; we evaluate it so that both updates are measured under both proposals).

\textbf{DIAL-MPC}~\cite{dialmpc} is distinct from the previous methods.
Its update is MPPI's with an adaptive temperature, \((\lambda = \tau \cdot \mathrm{std}_i(S^{(i)}))\), and an annealing schedule that shrinks the proposal noise across levels.
Each call of the annealer sequentially unrolls the sampler \(N\) times per replan, each level drawing a fresh batch of \(K\) rollouts across a horizon of \(H\) knots.
By default, SMC does not use an annealing schedule (\(N = 1\)) for a total of \(H \times K\) steps.
We include it for comparison in our MuJoCo tests and Push-T experiments.

In summary, the \(U_{\mathrm{mean}}\) and \(U_{\mathrm{elite}}\) updates average across many particles, whereas \(U_{\mathrm{MAP}}\) selects the single highest-weight trajectory and \(U_\mathrm{sample}\) takes a random sample.
With DIAL-MPC as the exception, \cref{alg:sisr} recovers each baseline using a single SIS/R procedure and modulating three axes.
In our experiments, we identify a method either by its string name, or equivalently its update rule and whether it resamples or not.

\begin{table*}[!t]
  \centering
  \begin{tabular}{lccccccccc}
    \toprule
    & \multicolumn{2}{c}{CartPole} & \multicolumn{2}{c}{DoubleCartPole} & \multicolumn{2}{c}{HalfCheetah} & \multicolumn{2}{c}{Walker} & \multirow{2}{*}{Norm.} \\
    \cmidrule(lr){2-3} \cmidrule(lr){4-5} \cmidrule(lr){6-7} \cmidrule(lr){8-9}
    Method & \multicolumn{1}{c}{Cost} & \multicolumn{1}{c}{ms} & \multicolumn{1}{c}{Cost} & \multicolumn{1}{c}{ms} & \multicolumn{1}{c}{Cost} & \multicolumn{1}{c}{ms} & \multicolumn{1}{c}{Cost} & \multicolumn{1}{c}{ms} & \\
    \midrule
    \multicolumn{10}{l}{\textit{Hydrax}~\cite{kurtz2024hydrax}} \\
    PS & 0.007 \(\pm\) 0.017 & 22.2  & 0.072 \(\pm\) 0.076 & 24.6  & 0.017 \(\pm\) 0.016 & 19.0  & 0.006 \(\pm\) 0.005 & 19.3  & 2.36 \\
    MPPI & 0.006 \(\pm\) 0.016 & 21.2  & 0.070 \(\pm\) 0.076 & 24.1  & 0.020 \(\pm\) 0.017 & 19.0  & \textbf{0.004 \(\pm\) 0.004} & 20.5  & 2.55 \\
    CEM & \textbf{0.005 \(\pm\) 0.014} & 20.7  & 0.102 \(\pm\) 0.073 & 23.4  & 0.043 \(\pm\) 0.009 & 19.6  & 0.007 \(\pm\) 0.008 & 22.4  & 4.70 \\
    DIAL-MPC & \textbf{0.005 \(\pm\) 0.014} & 99.5  & 0.150 \(\pm\) 0.038 & 109.6 & 0.041 \(\pm\) 0.012 & 96.7  & 0.009 \(\pm\) 0.017 & 93.7  & 4.82 \\
    \addlinespace[0.4em]
    \multicolumn{10}{l}{\textit{PFMP}} \\
    iCEM & 0.006 \(\pm\) 0.014 & 20.5  & \textbf{0.068 \(\pm\) 0.075} & 24.0  & 0.042 \(\pm\) 0.007 & 19.9  & 0.018 \(\pm\) 0.028 & 21.3  & 5.11 \\
    Biased-PS (PPO) & 0.006 \(\pm\) 0.015 & 24.2  & 0.098 \(\pm\) 0.074 & 30.5  & 0.004 \(\pm\) 0.009 & 29.8  & 0.053 \(\pm\) 0.029 & 25.0  & 4.08 \\
    Biased-MPPI (PPO) & 0.006 \(\pm\) 0.014 & 20.9  & 0.104 \(\pm\) 0.075 & 25.7  & \textbf{0.003 \(\pm\) 0.008} & 23.3  & 0.052 \(\pm\) 0.029 & 22.0  & 3.98 \\
    PS $(\rho = 0)$    & 0.006 \(\pm\) 0.016 & 21.2  & 0.080 \(\pm\) 0.076 & 22.5  & 0.005 \(\pm\) 0.008 & 20.0  & 0.005 \(\pm\) 0.003 & 20.7  & \textbf{1.28} \\
    PS $(\rho = 0.5)$  & 0.006 \(\pm\) 0.016 & 25.3  & 0.081 \(\pm\) 0.077 & 27.6  & 0.005 \(\pm\) 0.008 & 23.2  & 0.005 \(\pm\) 0.003 & 22.9  & \textbf{1.26} \\
    \bottomrule
  \end{tabular}
  \vspace{0.5em}
  \caption{Cost (mean \(\pm\) std) and per-replan wall time (ms), separated by implementation: the upper block is Hydrax, the lower is PFMP (our implementation).
    iCEM is our own subclass of Hydrax's CEM.
    \textbf{Norm.}\ is each method's mean cost across the four tasks normalized to the best method per task (lower is better; \(1.0\) = task winner), summarizing how consistently a method stays near the per-task optimum.
    The programmable proposals for Biased-PS and Biased-MPPI were learned offline with PPO; they never resample.
    DIAL-MPC was run with \(N = 5\) annealing iterations, which is why it runs around 5 times slower than others.
    Our training-free PS implementation (Gaussian proposal, \(\rho = 0\) or \(0.5\)) stays within \(1.3\times\) of the best on average.}
  \label{tab:mujoco}
\end{table*}

\section{Experiments}

Three experiment families explore the effect of varying two axes of \cref{tab:decomposition}, \ie, the resampling threshold (SIS \vs SIS/R) and the proposal (non-Gaussian mixtures \vs Gaussian baselines): kinematic bicycle navigation among obstacles, 7-DoF motion planning in the MotionBenchMaker (MBM) suite~\cite{chamzas2022motionbenchmaker}, and non-prehensile push-T among obstacles, simulated and on a real FR3.
We compare MPPI, Biased-MPPI, CEM, PS, and DIAL-MPC baselines with our own implementations.
As a regression test against Hydrax and DIAL-MPC implementations, we validate against four non-linear control tasks in MuJoCo MJX~\cite{todorov2012mujoco}, shown in \cref{tab:mujoco}.
Our implementation converges to good solutions over short horizons.

\subsection{2D Kinematic Bicycle}

\begin{figure*}[ht]
  \centering
  \includegraphics[height=5.0cm]{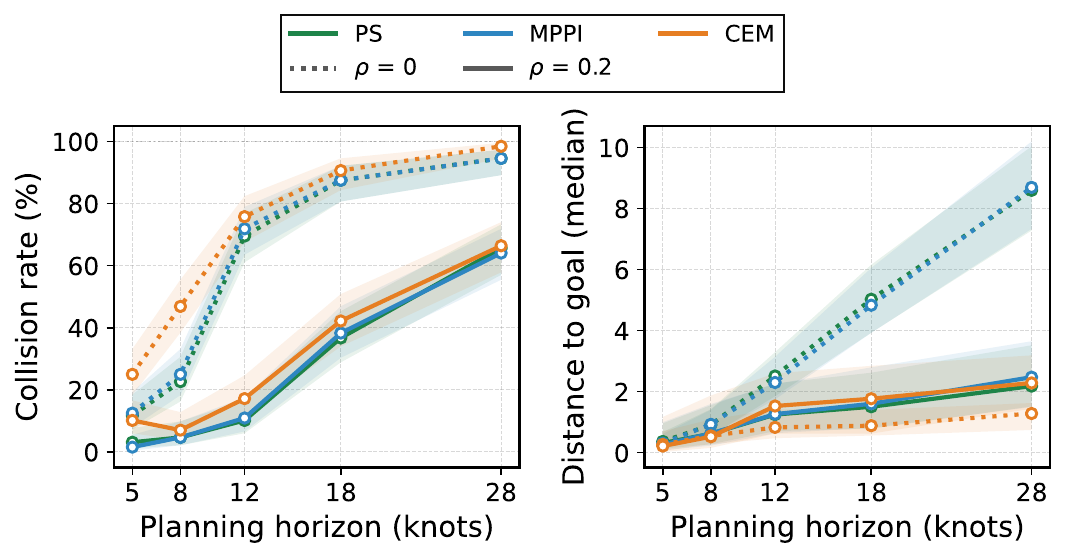}
  \hfill
  \includegraphics[height=5.0cm]{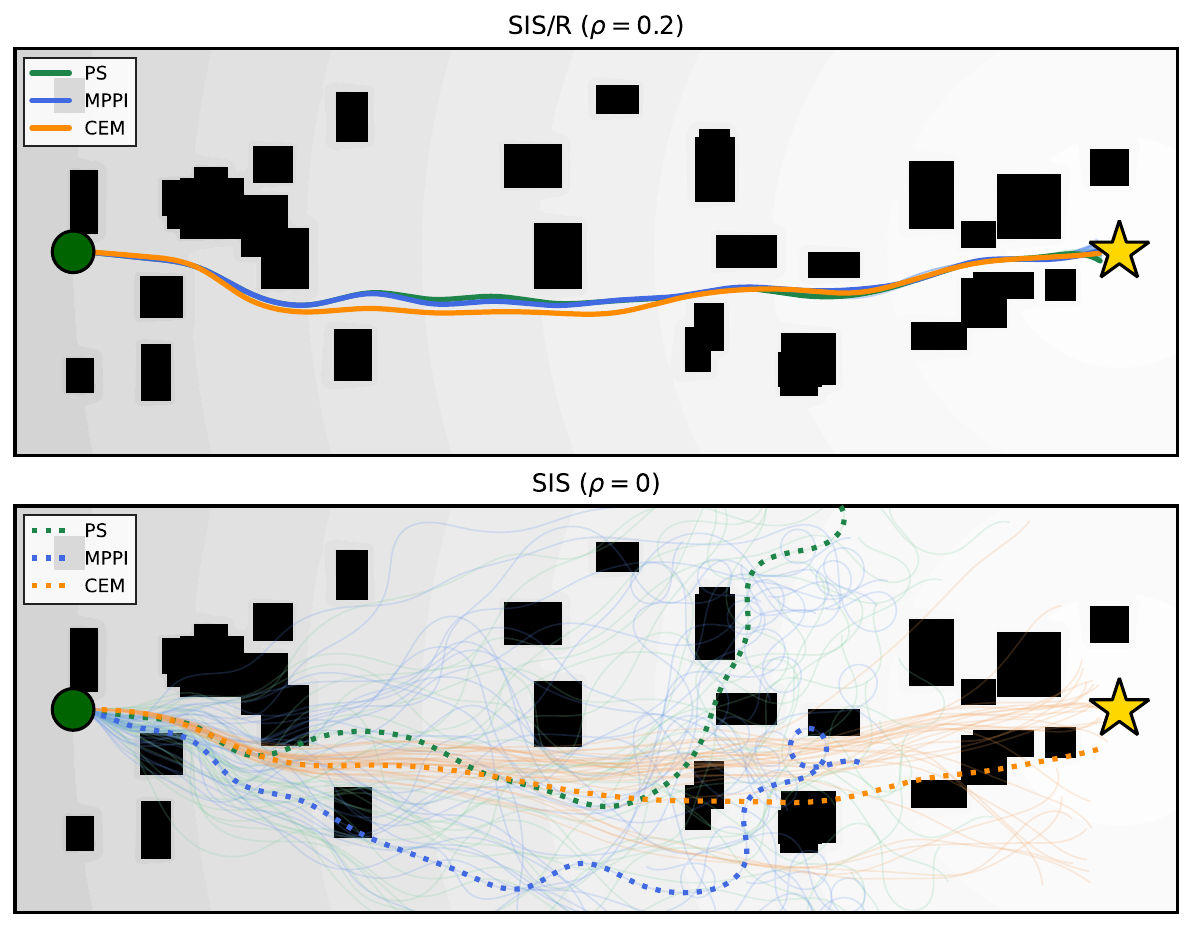}
  \caption{Open-loop planning through random AABB obstacle fields, 128 maps per horizon, \(K=1024\) particles.
    Every method receives the same budget: 20 refinement iterations of \(K\) particles over \(H\) knots.
    (\textit{Left}) Collision rate and median distance-to-goal against planning horizon.
    Hue is the update rule, a dotted line is \(\rho=0\) (SIS) and a solid line its resampling variant; the arena grows with the horizon.
    (\textit{Right}) The particle cloud and nominal plan at the final refinement pass on one map, with resampling (\textit{top}) and without it (\textit{bottom}).
    All methods use \(q_\mathrm{perturb}\).
    No trajectory post-processing or tuning is applied to any method, isolating the effect of the resampling axis.
  }
  \label{fig:2d-navigation}
\end{figure*}

Our first experiment was a 2D driving task, in which a kinematic bicycle agent must navigate through an AABB obstacle field.
We generated 128 maps at each of five horizons (\(H \in \{5, 8, 12, 18, 28\}\) knots) and scaled the arena with the horizon so that the problem difficulty effectively grows with the plan length.
In the method we only vary the resampling threshold \(\rho\) (SIS \vs SIS/R), and study its effect on planning in a 2D driving task.
Each method receives a finite budget of \(K = 1024\) particles and exactly 20 refinement iterations over \(H\) knots.
The plan for these experiments was constructed without changing the planning window; the goal is to study the planning algorithm without the effect of closed-loop replanning.

At the shortest horizon the different algorithms we tested are close.
As the horizon grows, SIS/R gains a clear advantage over SIS.
At 5 knots resampling takes the MAP update from a \(0.12\) collision rate to \(0.03\). At 28 knots the same collision rate went from \(0.95\) to \(0.66\), and drops the median distance-to-goal from \(8.6\) to \(2.2\).
Resampling prunes particles that would collide early, reallocating mass onto viable sequences.
\cref{fig:2d-navigation} (right) shows how without resampling the population scatters across the arena; despite the cost-weighting, non-resampling updates yielded collision-prone paths.
The elite update is instructive.
Non-resampling CEM drove most directly to the long-horizon (28 knot) goal, achieving the \textit{lowest} goal distance of \(1.3\), but without resampling (SIS), it collides in 98\% of maps.

\subsection{7-DoF Motion Planning}

Our second experiment is a 7-DoF motion planning task in which the Franka Research 3 (FR3) must plan a collision-free motion from a start configuration to the goal.
We evaluate on the MotionBenchMaker (MBM) benchmark suite, across seven distinct scene types with 100 start--goal problems each; a problem is solved if the executed plan is collision-free and $\lVert q_H - q_{\mathrm{goal}} \rVert_2 < 0.5$ radians.
Plans are five knots in length, and each method was given $K = 1000$ particles and 1 refinement iteration.
The programmable proposal here is a uniform three-way mixture over the goal configuration itself, a local Gaussian step with stdev \(\sigma\), and a uniform draw from the joint limits.
Averaged over seven datasets the programmable proposal increases success to 78--86\% (over Gaussian proposal baselines at 13--36\%). Resampling further increases success rate by 4 to 7 percentage points over the programmable proposal alone.

\begin{figure}[!t]
  \centering
  \includegraphics[width=\columnwidth]{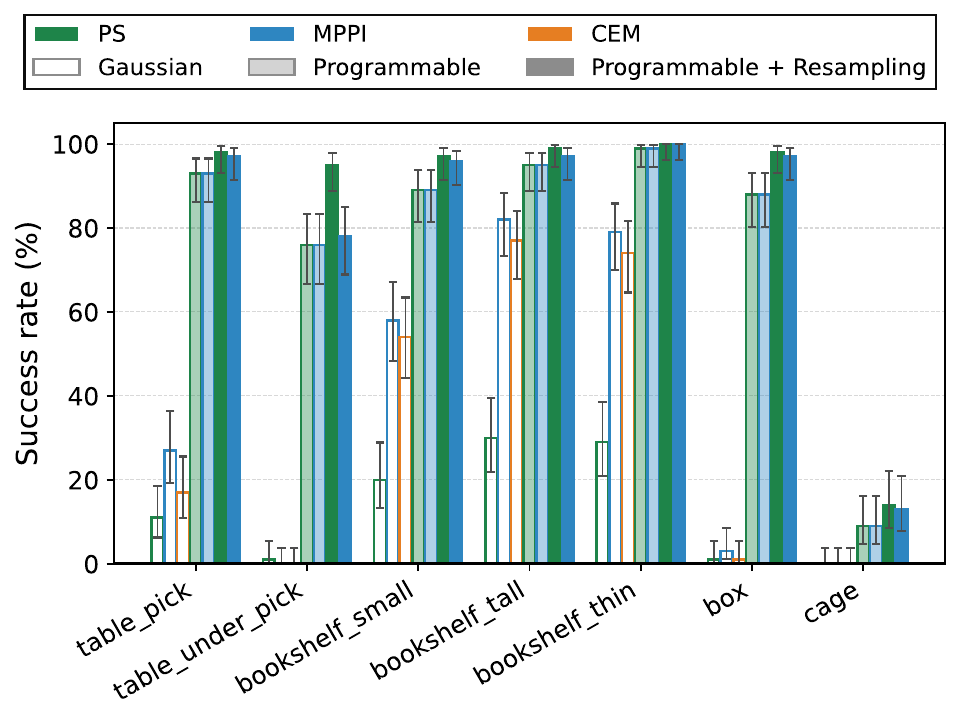}
  \caption{Success rate on MotionBenchMaker 7-DoF configuration-space planning: 100 problems per dataset, \(K=1000\) particles, error bars are 95\% Wilson intervals.
    Hue is the update rule; the Gaussian (transparent) and programmable proposals with $\rho = 0$ (translucent) are ablations of full SIS/R sampling (solid).
    Per-replan time is 50--100\,ms for all of them.}
  \label{fig:mbm}
  \vspace{-2em}
\end{figure}

\subsection{Closed-Loop Manipulation: Push-T in Simulation}

\begin{algorithm}[t]
  \caption{Non-prehensile Pushing Proposal Program.
    Mixture weights \((\alpha_1,\alpha_2,\alpha_3)=(0.4,0.2,0.4)\); steps toward the block are clipped to \(\delta=5\)cm; orbit radius \(r\) is fit to the object dimensions (0.8\(\times\) the width of the cap of the T).
    \(\mathbf{p}_T\) is the block's tracked position and \(\mathbf{c}_T\) the orbit center.
    \(\mathrm{clip}_\delta(v)=v\,\min(\lVert v\rVert,\delta)/\lVert v\rVert\).
  Line~\ref{line:project} applies on hardware only.}
  \label{alg:push_tee-proposal}
  \begin{algorithmic}[1]
    \Require waypoint \(u\in\mathbb{R}^2\); T position \(\mathbf{p}_T\), orbit center \(\mathbf{c}_T\)
    \State \(b \sim \mathcal{U}(0,1)\)
    \If{\(b < \alpha_1\)} \Comment{drive at the block}
    \State \(s \gets u + \mathrm{clip}_\delta(\mathbf{p}_T - u) + \sigma\eta,\quad \eta\sim\mathcal{N}(0,I_2)\)
    \ElsIf{\(b < \alpha_1 + \alpha_2\)} \Comment{reposition around the block}
    \State \(s \gets \mathbf{c}_T + r\,(\cos\varphi,\ \sin\varphi)\), \quad
    \(\varphi \sim \mathcal{U}(0,2\pi)\)
    \Else \Comment{local exploration}
    \State \(s \gets u + 2\sigma\eta,\quad \eta\sim\mathcal{N}(0,I_2)\)
    \EndIf
    \State \(s \gets \textsc{Project}(s,\ \mathcal{R})\) \label{line:project} \Comment{nearest IK-reachable cell}
    \State \textbf{return} \(s\)
  \end{algorithmic}
\end{algorithm}

\begin{figure*}[ht]
  \centering
  \includegraphics[height=6.0cm]{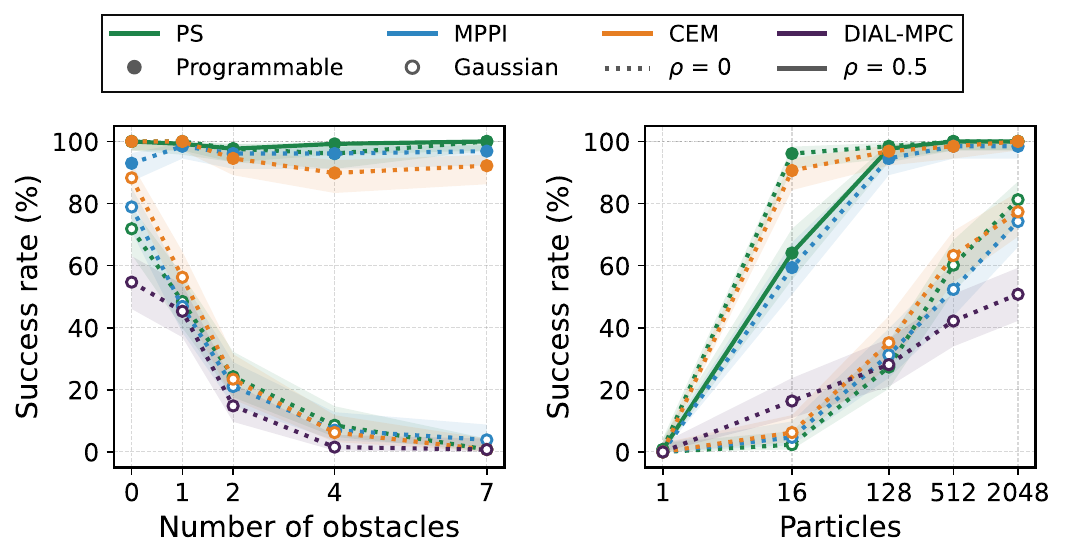}
  \hfill
  \includegraphics[height=6.0cm]{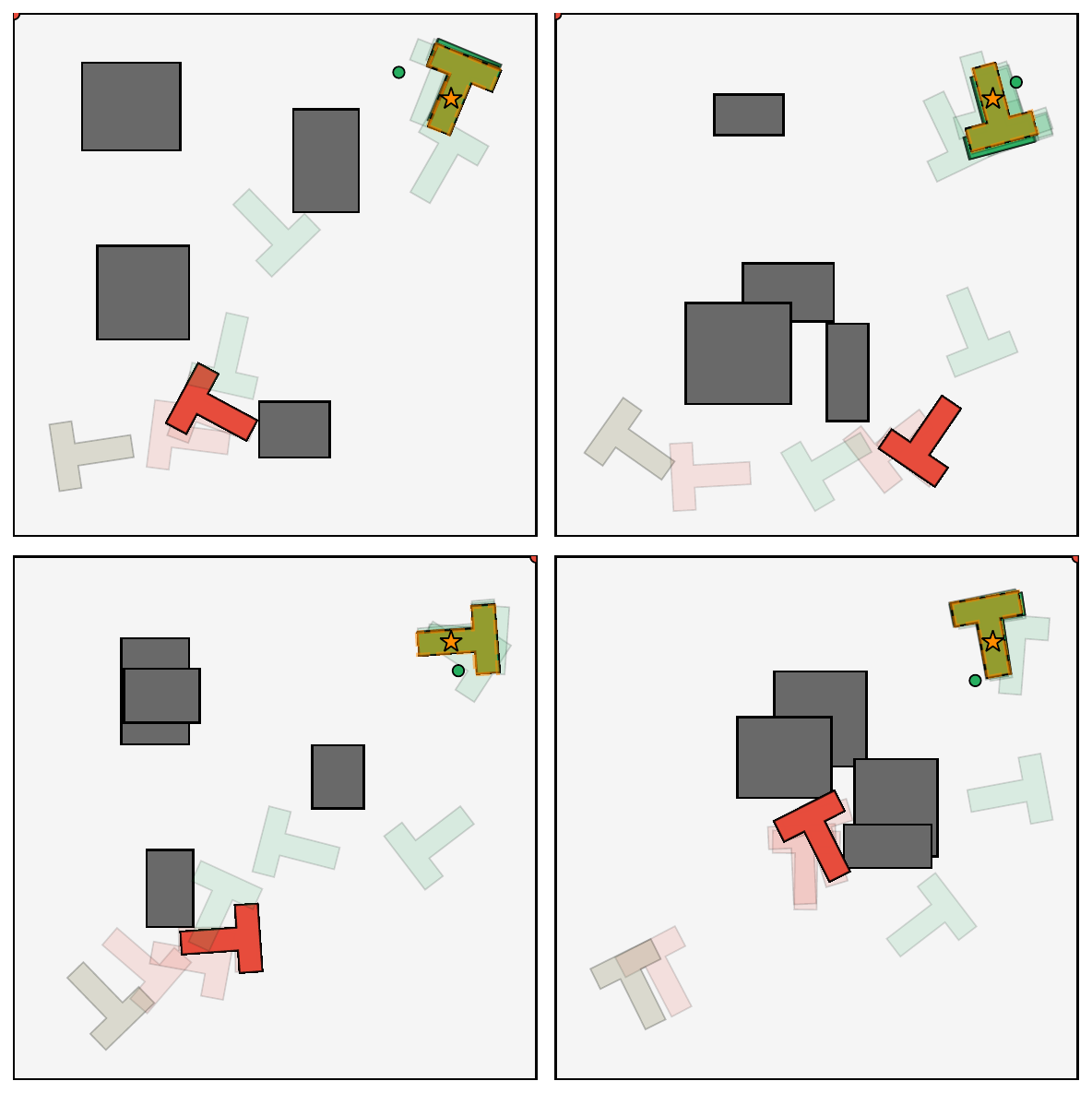}
  \caption{Push-T among obstacles. Every setting on the x-axis was tested across 128 randomly generated arenas. Shaded bands are 95\% Wilson intervals, and marker fill is the proposal, filled for programmable and hollow for Gaussian.
    (\textit{Left}) Varying the number of obstacles, at \(K=256\) particles.
    (\textit{Center}) Varying the particle count $K$, at a fixed one-obstacle difficulty.
    (\textit{Right}) Trajectory traces for PS with Gaussian proposal (red) and programmable proposal (green) on four scenarios with four obstacles.
    The programmable proposal keeps the end-effector engaged with the block, while the perturbative baseline drifts away from it.
  T-block poses are drawn at each replanning step, with the goal pose in orange.}
  \label{fig:push-t}
\end{figure*}

Push-T is a standard non-prehensile manipulation benchmark popularized by
implicit behavioral cloning~\cite{florence2022implicit} and diffusion
policy~\cite{chi2023diffusion}, in which an impedance-controlled circular robot must push a T to a goal configuration.
In order to highlight the effect of having a programmable proposal, we consider a more difficult variant, where the T must be navigated around randomly sampled AABB obstacles.
For our trials, we consider it a success if the block comes within \(5\)\,cm Euclidean distance and \(0.2\)\,rad of the goal pose.

The SMC proposal is the three-branch mixture of \cref{alg:push_tee-proposal}: a step toward the T block, an orbital branch that repositions the end-effector around it, and a local random walk.
Each method was tuned using Optuna on a disjoint set of problems.
We tuned the sampling-noise width \(\sigma\) for every method; the rest of the tuning
space is determined by the hyperparameters used by each update rule $U$; we tune temperature \(\lambda\) for the MPPI
variants, elite count and covariance floor for the CEM variants, and the
annealing parameters for DIAL-MPC. The mixture probabilities $\alpha_1$, $\alpha_2$, $\alpha_3 \in [0, 1]$ in our programmable proposal are held fixed. We run two experimental sweeps.

\textbf{Varying Problem Difficulty.} We hold particle count fixed at $K = 256$ and generate arenas with varying number of obstacles (\{0, 1, 2, 4, 7\}), which lets us explore the effect of the proposal as problem difficulty increases.
Every method solves a majority of arenas when there are no obstacles, but the Gaussian group (PS 72\%, MPPI 79\%, CEM 88\%, DIAL-MPC 55\%) already trails programmable proposals which achieve 100\%.
The Gaussian-proposal group totally collapses as obstacles are added: at seven obstacles, PS, CEM, and DIAL-MPC are all near zero. Plans remain consistently better under the programmable proposal: Biased-PS at 100\%, Biased-MPPI at 97\%, and Biased-CEM at 92\%.
At this difficulty, the proposal is the dominant factor influencing success, raising the resulting success rate by 90\% across the board, while the update rule is negligible.

\textbf{Varying Particle Budget.} We hold the arena size fixed with one obstacle, and vary the particle budget $K$ in \{1, 16, 128, 512, 2048\}.
The group with the programmable proposal reaches 96\% (Biased-PS) and 91\% (Biased-CEM), with 16 particles.
The best baseline method achieves a lower success rate of 81\% (PS), with 2048 particles.
The programmable proposal in this experiment not only improves success rate, it does so with two orders of magnitude fewer particles.

\subsection{Closed-Loop Manipulation: Push-T on Hardware}
\label{sec:franka}

This final experiment evaluates the performance of~\cref{alg:push_tee-proposal} on a real-world robot.
We reused the proposal from the simulated push-T task on the FR3, with only a single addition: the planar end-effector waypoints must be reachable by inverse kinematics (IK).
At startup we memoize IK solutions using the GeoFIK solver~\cite{lopezcustodio2025geofik} over a \(128 \times 128\) grid of the workspace, for \((x,y)\) translations at a fixed height of \(6\)\,cm with the end-effector pointing down.
Cells admitting a solution within the workspace bounds (\(r \geq 0.29\)\,m,
\(|y| \leq 0.78\)\,m) form the reachable set \(\mathcal{R} \subset \mathbb{R}^2\),
which is a partial annulus centered in front of the robot.
Both the programmable and the Gaussian proposals sample waypoints \(s \in
\mathbb{R}^2\) and then apply \(\textsc{Project}(s, \mathcal{R})\), which snaps
\(s\) to the nearest cell of \(\mathcal{R}\).
The dynamics function indexes into the memoized IK joint configuration, and takes a step toward the desired configuration using MuJoCo's PD solver.
The running cost function for the real task contains one additional term:
the arm's reachable set is an annulus, so we add a radial penalty \(w_a(\lVert \mathbf{p}_T \rVert - r_a)^2\) on the block's distance from the annulus midpoint \(r_a = 0.549\)\,m, incentivizing the object to be closer to the center of the reachable region.

State estimation uses an OptiTrack motion capture system and the T-block geometry is
loaded from CAD; \cref{fig:fr3-setup} shows the setup.
The goal is reached if the T-block's pose comes within 5cm Euclidean distance and $20^\circ$ of the goal pose in under 90 seconds.
We evaluate each method across 20 identical (start, goal) trial pairs, generated deterministically from a fixed seed. For reproducibility, we matched the specified start pose for each method to within 3.5mm on average.

Biased-PS with the proposal program~\cref{alg:push_tee-proposal} achieves 90\% success rate, while the Gaussian baselines make little progress: PS, MPPI, and DIAL-MPC each reach 10\% and CEM 15\%.
Biased-MPPI, which shares a proposal but applies the MPPI update, reaches 75\%.

\section{Discussion}

The current work has a few limitations.
The set of proposals that we explore is relatively small; future work on scheduling, learning proposals, or amortized neural inference may offer better coverage of the posterior.
Our state estimation relies on an accurate motion capture system and fast replanning.
Furthermore, we assume access to a relatively accurate dynamics model with known features such as the tabletop dimensions and T shape modeled directly from known 3D geometries.
MuJoCo's contact model occasionally shows large instabilities and overall has a poor representation of real-world sliding and torsional friction effects.
Nevertheless, replanning is what makes MPC robust to residual simulator error.
We consider planning with state uncertainty rather than around it to be an important extension.

One significant empirical finding: in Push-T with one obstacle, resampling \emph{hurts} for 16 particles. SIS/R reaches 64\% against SIS's 96\%, showing that resampling is not universally advantageous for planning.
Careful tempering schedules may maintain better particle diversity in the more challenging cost landscapes presented by contact-rich tasks.

Further study of SMC for plan inference in C-space environments, including formalization of its properties under different cost models, could be a valuable refinement of this work.
Our SMC decomposition takes one step toward explaining and improving sampling-based MPC stacks with concepts from algorithmic inference.
Through our ablation experiments, we find that it's not the update rule, but rather the programmable proposal and resampling that make the biggest difference for MPC.

\printbibliography

\end{document}